\pdfoutput=1

\documentclass[11pt]{article}

\usepackage[preprint]{emnlp}

\usepackage{times}
\usepackage{latexsym}

\usepackage[T1]{fontenc}

\usepackage[utf8]{inputenc}

\usepackage{microtype}

\usepackage{inconsolata}

\usepackage{graphicx}

\usepackage{multirow}
\usepackage{adjustbox}
\usepackage{booktabs}
\usepackage{makecell}
\usepackage{float}
\usepackage{tabularx}
\usepackage[most]{tcolorbox}
\usepackage{graphicx}
\usepackage{subcaption}

\title{Scaling Personality Control in LLMs with Big Five Scaler Prompts}

\newcommand\correspondingauthor{\thanks{Corresponding author.}}

\author{
 \textbf{Gunhee Cho\textsuperscript{1}},
 \textbf{Yun-Gyung Cheong\correspondingauthor\textsuperscript{1}}
\\
 \textsuperscript{1} Sungkyunkwan University, South Korea
\\
  \textsuperscript{1} \{skate4333, aimecca\}@skku.edu
}

\begin{document}
\maketitle
\begin{abstract}
We present \textit{Big5-Scaler}, a prompt-based framework for conditioning large language models (LLMs) with controllable Big Five personality traits. By embedding numeric trait values into natural language prompts, our method enables fine-grained personality control without additional training. We evaluate Big5-Scaler across trait expression, dialogue generation, and human trait imitation tasks. Results show that it induces consistent and distinguishable personality traits across models, with performance varying by prompt type and scale. Our analysis highlights the effectiveness of concise prompts and lower trait intensities, providing a efficient approach for building personality-aware dialogue agents.

\end{abstract}

\section{Introduction}
Equipping large language models (LLMs) with distinct and controllable personalities is an emerging goal in dialogue research, aimed at improving user engagement, consistency, and social alignment. 
As LLMs are increasingly deployed in applications that involve direct interaction with end users, such as conversational agents and educational tutors, there is a growing need for methods that enable dynamic persona control while minimizing resource costs \citep{frisch2024llm, achiam2023gpt}. Prior work typically relies on curated character data or persona-specific fine-tuning, which limits scalability across diverse use cases \citep{zhang2018personalizing, roller2021recipes}.

Prior work on persona agents often relies on curated character data, such as dialogue transcripts, profile descriptions, or biographies, to inject personality into language models \citep{zhang2018personalizing, majumder2020like, wang2024rolellmbenchmarkingelicitingenhancing}.  
These approaches typically involve fine-tuning or few-shot prompting using character-specific inputs.  
 While effective in controlled settings, they require substantial manual curation, domain expertise, and computational resources, limiting their scalability across diverse persona types \citep{roller2021recipes}.


Furthermore, existing methods offer limited flexibility in modulating the intensity of personality expression. For instance, a model fine-tuned on a cheerful persona may consistently adopt an upbeat tone but cannot adjust the degree of expressiveness without retraining. This constraint hinders the dynamic generation of persona agents with nuanced or composite traits for interactive and adaptive applications \citep{jiang2024personallminvestigatingabilitylarge, jiang2023evaluatinginducingpersonalitypretrained}.

To address these challenges, we introduce \textit{Big5-Scaler}, a prompt-based approach to persona conditioning grounded in the Big Five personality theory \citep{big5}. Our method assigns explicit numeric values (e.g., 0–100) to each trait dimension—Openness, Conscientiousness, Extraversion, Agreeableness, and Neuroticism—enabling fine-grained control over the degree of trait expression. These values are encoded in LLM prompts provided at inference time, removing the need for fine-tuning on persona-specific training data.

Our empirical evaluation indicates that the proposed method reliably elicits personality-consistent behavior in LLM agents as evidenced by average Big Five trait scores exceeding 4.0 (max 5, min 1) and a PersonaCLR score of approximately 0.8 (max 1, min 0). The generated dialogues reflect the intended trait intensities, demonstrating that the approach supports scalable and flexible generation of diverse persona agents. 

This work contributes a efficient, controllable, and training-free framework for persona 
construction. In future agent simulation environments \citep{park2023generativeagentsinteractivesimulacra, park2024generativeagentsimulations1000}, the Big5-Scaler could be utilized to efficiently assign personality profiles to agents, enabling rapid simulation setup. Beyond that, one could envision the use of agent-based simulations to empirically explore psychological hypotheses, such as romantic compatibility based on Big Five personality traits. \citep{WEIDMANN2023104378}


Our work makes the following contributions:
\begin{itemize}
    \item We propose \textit{Big5-Scaler}, a prompt-based framework for personality conditioning that requires no curated character data or fine-tuning. The method leverages the Big Five personality theory to induce interpretable and controllable personality traits in LLMs.
    
    \item We introduce a trait-level control mechanism that assigns explicit numerical values to each Big Five dimension, enabling fine-grained modulation of personality expression directly at inference time.
    
    \item We conduct comprehensive empirical evaluations—including trait expression analysis, dialogue-based assessment, and human imitation—showing that the generated outputs consistently reflect the specified trait configurations across models and tasks.
\end{itemize}



\begin{figure*}[t]
    \centering
    \includegraphics[width=1\textwidth]{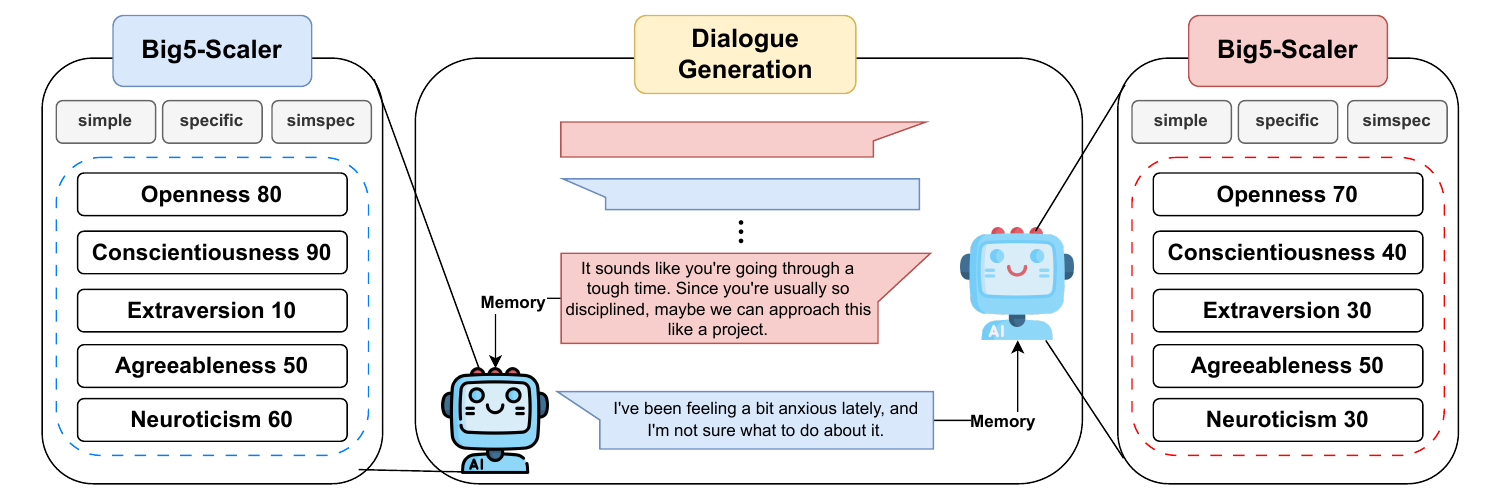}
    \caption{Overview of the \textbf{Big5-Scaler} method. Persona prompts are constructed using one of three prompt types (simple, specific, or simspec) and assigned to agents based on Big Five trait values. In this example, the scale \textit{n} is set to 100. Two agents with different trait configurations engage in dialogue, during which their utterances are generated and stored in each agent's memory.}
    \label{fig:big5-scaler_method}
\end{figure*}

\section{Related Work}
A growing body of research has explored methods for endowing LLMs with stable, interpretable, and controllable personality traits to enable more coherent and engaging interaction. This section reviews prior work on LLM-based personality simulation and the integration of Big Five personality theory into natural language processing.

\subsection{Simulating Human-Like Personality}
Recent work has investigated whether LLMs can consistently and interpretably simulate stable personality traits across multi-turn interactions \citep{frisch2024llm}.
To evaluate these capabilities more systematically, TRAIT was introduced as a large-scale benchmark comprising over 8,000 multiple-choice questions derived from validated psychometric instruments such as the Big Five Inventory (BFI) and the Short Dark Triad (SD-3) \citep{lee2025llmsdistinctconsistentpersonality}. Results suggest that LLMs exhibit distinct and persistent personality profiles that are influenced by both pretraining and alignment processes.

Complementary work has developed methodologies for administering personality assessments to LLMs, showing that instruction-tuned models more reliably simulate psychologically meaningful traits \citep{serapiogarcía2025personalitytraitslargelanguage}. For instance, GPT-3.5 and GPT-4 have been shown to align with Big Five traits across self-report inventories and narrative generation tasks, highlighting the efficacy of trait-level prompting \citep{jiang2023evaluatinginducingpersonalitypretrained}. Further, interview-style prompts have been used to elicit trait-consistent behavior, revealing that LLMs modulate personality expression based on input phrasing \citep{hilliard2024elicitingpersonalitytraitslarge}.

In parallel, several persona-guided approaches have explored the use of narrative-derived character data to shape personality expression. These include methods that fine-tune models on raw scripts or dialogues \citep{shao-etal-2023-character}, prompt models with character-specific descriptions \citep{zhou2023characterglmcustomizingchineseconversational}, or directly train on structured persona attributes \citep{yu2024neekoleveragingdynamiclora}. More elaborate settings incorporate detailed character profiles to produce agents with rich, distinctive personalities \citep{li2023chatharuhirevivinganimecharacter, wang2024rolellmbenchmarkingelicitingenhancing}.

\subsection{Modeling Big Five Personality}
The Big Five personality theory—comprising Openness, Conscientiousness, Extraversion, Agreeableness, and Neuroticism—has long served as a foundational framework in personality psychology \citep{big5}. To assess individual trait levels, psychologists have developed standardized instruments such as the Big Five Inventory (BFI) \citep{john1991big}, IPIP-NEO \citep{ipip-neo}, IPIP-NEO-120 \citep{ipip-neo-120}, NEO PI-R \citep{neo-pi-r}, and NEO-FFI \citep{costa2008revised}. More recent efforts have extended these frameworks to finer-grained assessments, such as the Multidimensional Personality Inventory (MPI), enabling more nuanced evaluation of personality traits \citep{jiang2023evaluatinginducingpersonalitypretrained}.


The Big Five model has been widely adopted for applications including personality recognition from text \citep{yeo-etal-2025-pado}, social media analysis \citep{10.1371/journal.pone.0303770, inbook}, and persona-grounded dialogue generation \citep{han2024psydialpersonalitybasedsyntheticdialogue, miyama2022personalityadaptedmultimodaldialogue}. More recent work focuses on equipping language models with the ability to internalize and express Big Five traits. For example, BIG5-CHAT presents a modular framework that trains expert components for each trait using synthetic dialogue data \citep{li2025big5chatshapingllmpersonalities}. 
P-Tailor employs a Mixture-of-Experts (MoE) architecture with trait-specific LoRA adapters, enabling modular and controllable personality simulation \citep{dan2024ptailorcustomizingpersonalitytraits}.

\section{Method}
\label{sec:method}
We present Big5-Scaler, a prompt-based framework for controlling personality expression in large language models by specifying the intensity of each Big Five trait.
Section \hyperref[sec:big5traits]{3.1} provides an overview of the Big Five personality dimensions.
Section~\ref{sec:big5scaler} details the design of the Big5-Scaler prompt format.
Section~\ref{sec:big5scaleragent} describes how agents are constructed and deployed using these prompts.
An overview of the system is illustrated in Figure~\ref{fig:big5-scaler_method}.

\subsection{Big Five Traits}
\label{sec:big5traits}
The Big Five personality theory defines human personality along five core dimensions: Openness (O), Conscientiousness (C), Extraversion (E), Agreeableness (A), and Neuroticism (N). This model is widely used in psychology to capture individual differences.

Each trait is defined as follows:
\begin{itemize}
\item\textbf{Openness}: Imaginative, curious, open to new experiences, and intellectually engaged. 
\item\textbf{Conscientiousness}: Organized, self-disciplined, goal-oriented, and reliable. 
\item\textbf{Extraversion}: Outgoing, energetic, sociable, and assertive.
\item\textbf{Agreeableness}: Compassionate, cooperative, trusting, and kind.
\item\textbf{Neuroticism}: Prone to negative emotions such as anxiety, anger, or depression.
\end{itemize}
Each trait is further subdivided into six facets, as defined by the Revised NEO Personality Inventory (NEO PI-R) \citep{neo-pi-r}. 
This hierarchical structure allows for a more fine-grained assessment of individual personality profiles. For detailed descriptions of each facet, please refer to the Appendix \hyperref[sec:big5_faucet_appendix]{A}.



\subsection{Big5-Scaler}
\label{sec:big5scaler}
We define \textbf{Big5-Scaler} as a personality conditioning method that encodes trait intensity values directly into natural language prompts. 
For each trait, the prompt includes (1) a definition, (2) a behavioral description, and (3) an assigned numerical value. The trait intensity scale $n$ is configurable; we use four discrete levels: 10, 25, 50, and 100.

We define three types of prompts:
\begin{itemize}
\item \textbf{Simple Prompt} : High-level descriptions of the five traits.
\item \textbf{Specific Prompt} : Facet-level behavioral descriptions for each trait.
\item \textbf{Simspec Prompt} : A combination of both trait-level and facet-level descriptions.
\end{itemize}
Examples of each prompt type are provided in Appendix~\ref{sec:big5_scaler_prompts_appendix}.

\subsection{Big5-Scaler Agent Construction}
\label{sec:big5scaleragent}
We construct a set of agents \( \mathcal{A} = \{ a_1, a_2, \dots, a_n \} \), where each agent \( a_i \) (where \( 1 \leq i \leq n \)) is initialized with a personality prompt \( p_i \) generated using Big5-Scaler. The agent architecture follows memory-based dialogue frameworks proposed in prior work \citep{chu2024cohesiveconversationsenhancingauthenticity, wang2024recursivelysummarizingenableslongterm}.


Each agent maintains a memory buffer \( M_i \), initialized with its personality prompt:
\begin{equation}
M_i = \{ p_i \}
\end{equation}

This setup can be extended to \(n\) agents. In the scenario described below, we consider the case of two agents for illustrative purposes. These two agents engage in turn-based dialogue. At turn \( j \), agent \( a_1 \) generates an utterance \( m_j \) conditioned on its current memory:
\begin{equation}
m_j = \text{Generate}(M_1)
\end{equation}
This message is appended to both agents' memory buffers:
\begin{equation}
M_1 \leftarrow M_1 \cup \{ m_j \}, \quad M_2 \leftarrow M_2 \cup \{ m_j \}
\end{equation}
The agents take turns generating utterances. At turn \( j+1 \), the responding agent generates the next utterance \( m_{j+1} \) based on its updated memory:
\begin{equation}
m_{j+1} = \text{Generate}(M_2)
\end{equation}
Again, both memories are updated:
\begin{equation}
M_1 \leftarrow M_1 \cup \{m_{j+1}\}, \quad M_2 \leftarrow M_2 \cup \{m_{j+1}\}
\end{equation}

This process continues iteratively, enabling agents to maintain distinct personality conditioning while responding to a shared dialogue history.





\section{Experiments}
We empirically evaluate the Big5-Scaler framework across multiple settings to evaluate the effectiveness of Big5-Scaler in controlling and simulating personality traits in LLMs.

We begin by evaluating whether Big5-Scaler prompts enable LLMs to generate text that clearly reflects the intended Big Five traits.  
Next, we examine whether the numerical intensity values assigned to each trait are accurately manifested in the model’s outputs.
We then assess trait consistency by measuring how well each personality configuration is maintained throughout multi-turn dialogues.
We further simulate dialogues between agents conditioned with different trait configurations using Big5-Scaler, and analyze how these trait differences influence the generated interactions. 
Finally, we investigate Big5-Scaler’s ability to mimic human personalities by prompting agents with human-assigned trait scores and comparing their behavioral outputs.


We evaluate our method using three high-performing open-source LLMs: LLaMA3-8B \citep{grattafiori2024llama3herdmodels}, Mistral-25B \citep{jiang2023mistral7b}, and Phi4-14B \citep{abdin2024phi4technicalreport}. As described in Section~\ref{sec:method}, we experiment with three Big5-Scaler prompt types: \textit{simple}, \textit{specific}, and \textit{simspec}. The generation was conducted with the settings of \texttt{max\_new\_tokens = 512}, \texttt{temperature = 1.0}, and \texttt{top\_p = 0.8}. These settings were chosen because persona agent generation requires a certain level of creativity, and overly restrictive decoding parameters may negatively affect performance.
Model-specific settings and evaluation metrics are detailed in the respective experimental sections.

\subsection{Single-Trait Expression Ability}
In this experiment, we evaluate the model's ability to express a specific Big Five trait when explicitly instructed to maximize that trait. 
We compare our approach against the \textbf{NEUTRAL} setting (no personality prompt), along with three prompting strategies proposed by \citet{jiang2023evaluatinginducingpersonalitypretrained}:

\textbf{NAIVE Prompting} \citep{brown2020language}:  
The model is prompted with a simple natural language instruction in the form of ``You are a/an X person,'' where \(X\) corresponds to one of the Big Five traits.

\textbf{WORDS AUTO Prompting}:  
We adopt a prompt search strategy inspired by prior work \citep{prasad2023gripsgradientfreeeditbasedinstruction, shin2020autopromptelicitingknowledgelanguage}.  
For each Big Five trait, we select the three most representative words from the candidate sets provided by \citet{KWANTES2016229}.  
We then evaluate personality expression using the BFI-S questionnaire \citep{bfi-s}.

\textbf{P2} \citep{jiang2023evaluatinginducingpersonalitypretrained}:  
This strategy improves upon naive prompting by first selecting trait-relevant keywords and then generating descriptive phrases that elaborate on those traits.  
This two-step process is designed to more effectively elicit the target personality dimension.

For our method, \textbf{Big5-Scaler}, we apply all three prompt types—\textit{simple}, \textit{specific}, and \textit{simspec}—using a fixed trait intensity score of 100 with scale \(n = 100\). In each trait-specific condition, only the target trait is described in the prompt, while the remaining traits are omitted. This setup ensures a fair comparison with baselines that target a single trait at a time.

All prompt-based evaluations are conducted using the Alpaca-7B model,\footnote{\url{https://github.com/tatsu-lab/stanford_alpaca}} which was shown to exhibit high consistency in personality expression in prior work \citep{jiang2023evaluatinginducingpersonalitypretrained}.  
We use the 1,000-item Machine Personality Inventory (MPI) dataset introduced in the same study as our evaluation benchmark.

We report the mean and variance of trait scores computed from the MPI questionnaire. Each Big Five trait score ranges from 1 to 5. A higher mean indicates stronger expression of the target trait, while a lower variance reflects greater consistency and robustness across samples.

\subsection{Proportional Trait Scaling Analysis}
To test whether higher assigned trait scores result in stronger expression of the corresponding personality trait in generated outputs, we conduct a trait tendency evaluation.

We apply three types of Big5-Scaler prompts (\textit{simple}, \textit{specific}, and \textit{simspec}) to several open-source LLMs. 
To evaluate the resulting outputs, we use two standardized Big Five personality questionnaires: BFI \citep{john1991big}, and IPIP-NEO-120 \citep{ipip-neo-120}.


In all experiments, the trait intensity scale is fixed at \(n = 100\). For each test, we vary the target trait across ten levels: 0, 10, 20, ..., 90, while holding the other four traits constant at a neutral value of 50. Each prompt configuration is used to instantiate an agent, which then completes the designated questionnaire. The score obtained for the target trait is recorded.

To evaluate proportionality, we plot the assigned trait value on the \(x\)-axis and the corresponding questionnaire-derived score on the \(y\)-axis. A strong linear relationship indicates that the numerical conditioning is effectively realized.

We quantify this relationship using the Pearson correlation coefficient \citep{pearson_vii_1895}. A coefficient close to 1 indicates strong positive correlation, while a low \(p\)-value suggests statistical significance and suggests that the observed relationship is unlikely to have occurred by chance.







\subsection{Trait Expression in Interactive Dialogue}
\label{sec:agent_dialogue}

We simulate dialogues between two agents, each initialized using Big5-Scaler with a distinct Big Five personality profile.
Prompt type and trait scale \( n \) vary across models. For LLaMA3-8B, we use the \textit{specific} prompt with \( n = 10 \), whereas for Mistral-25B and Phi4-14B, we apply the \textit{simple} prompt with the same scale. These settings are informed by an analysis (Section~\ref{sec:analysis}), which showed that they most accurately reflect trait intensity.

Each dialogue consists of 20 turns, with each agent contributing 10 utterances. Dialogue topics are randomly selected from a predefined set: \textit{travel}, \textit{music}, \textit{habits}, \textit{goals}, \textit{friends}, \textit{social events}, \textit{animals}, \textit{volunteering}, \textit{self-esteem}, and \textit{anxiety}.

Prior to each dialogue, agents are randomly assigned trait values.  
Once the conversation is generated, we evaluate the resulting dialogues using external LLM-based evaluators: GPT-4o-mini \citep{achiam2023gpt}, Claude 3.5 Haiku,\footnote{\url{https://www.anthropic.com/claude/haiku}} and DeepSeek-Chat \citep{guo2025deepseek}.

\subsection{Intra-Dialogue Trait Consistency}
This experiment evaluates whether Big5-Scaler agents maintain consistent personality expression over the course of a dialogue.
Following the PersonaCLR framework \citep{personaclr}, we extract 10 utterances from each agent's dialogue (generated in Section~\ref{sec:agent_dialogue}). 
We concatenate the first nine utterances and measure their similarity to the final (tenth) utterance.  
A higher similarity score indicates greater consistency in personality expression across turns.

We use three evaluation methods: cosine similarity \citep{salton1975vector}, Sentence-BERT embeddings \citep{reimers-gurevych-2019-sentence}, and the PersonaCLR metric \citep{personaclr}. The model and Big5-Scaler prompt settings used for the agents are identical to those described in Section~\ref{sec:agent_dialogue}.


We adapted the original PersonaCLR setup to support English.  
The original study used Waseda University’s Japanese RoBERTa model,\footnote{\url{https://huggingface.co/nlp-waseda/roberta-large-japanese-with-auto-jumanpp}} fine-tuned on the Naro Utterance (NaroU) dataset \citep{personaclr}.  
In contrast, we employed the multilingual \texttt{xlm-roberta-base} model,\footnote{\url{https://huggingface.co/FacebookAI/xlm-roberta-base}} trained on the same dataset.  
This modification enables PersonaCLR to be applied to English, thereby extending its usability beyond Japanese.


\subsection{Human-to-Agent Trait Alignment}

A key strength of Big5-Scaler is its ability to generate agents that reflect real human personality profiles.  
This is achieved by directly mapping human Big Five scores into Big5-Scaler prompts.

To evaluate this capability, we conducted a study with 17 Korean participants, all graduate students conducting research in natural language processing.  
Each participant completed the IPIP-NEO-120 questionnaire, and their Big Five trait scores were used to construct corresponding Big5-Scaler prompts.  
These prompts were then used to generate personality-aligned agents.


Each generated agent was then administered the IPIP-NEO-120 questionnaire.  
We computed the root mean squared error (RMSE) between the agent's trait scores and those of the corresponding human participant.  
Lower RMSE values indicate closer alignment, reflecting the agent's ability to accurately mimic the target personality profile.

\section{Results}

\subsection{Single-Trait Expression Ability}

Overall, models guided by Big5-Scaler exhibit stronger expression of target traits compared to baseline methods, as reflected by higher mean scores across most dimensions (Table~\ref{tab:big5-induce}).

One consistent exception is the Neuroticism trait, where all methods, including Big5-Scaler, underperform. A likely explanation is that neurotic traits involve negative affective states such as anxiety or anger, which LLMs are typically discouraged from expressing due to safety alignment objectives. Since most large language models are trained to avoid toxic, emotionally unstable, or harmful content, the controlled simulation of neurotic behaviors may be inherently constrained.

In terms of robustness, Big5-Scaler achieves comparable variance to baseline methods.  
Notably, unlike WORDS and P2 which rely on preprocessing steps to construct trait-specific prompts, Big5-Scaler operates without any auxiliary procedures.  
This highlights its efficiency, as it achieves similar or better consistency with a simpler pipeline.

Together, these results demonstrate that Big5-Scaler provides an efficient and effective prompting strategy for inducing personality-aligned behavior in LLMs, without requiring task-specific tuning or data construction.





\begin{table*}[t]
\begin{adjustbox}{width=0.8\textwidth,center}
\begin{tabular}{c|cc|cc|cc|cc|cc}
\toprule
\multirow{2}{*}{\textbf{Model}} & \multicolumn{2}{c|}{\textbf{O}{\scriptsize penness}} & \multicolumn{2}{c|}{\textbf{C}{\scriptsize onscientiousness}} & \multicolumn{2}{c|}{\textbf{E}{\scriptsize xtraversion}} & \multicolumn{2}{c|}{\textbf{A}{\scriptsize greeableness}} & \multicolumn{2}{c}{\textbf{N}{\scriptsize euroticism}} \\
 & Score & $\sigma$ & Score & $\sigma$ & Score & $\sigma$ & Score & $\sigma$ & Score & $\sigma$ \\
\midrule
Neutral & 3.97 & 1.41 & 3.61 & 1.56 & 3.89 & 1.51 & 3.56 & 1.64 & \textbf{3.01} & 1.65 \\
\midrule
NAIVE   & 3.17 & 1.29 & 3.19 & 1.11 & 2.85 & 1.06 & 3.11 & 1.05 & \underline{2.83} & 1.30 \\
WORDS   & 3.53 & 1.25 & 3.12 & 1.09 & 3.03 & 1.09 & 3.33 & 1.11 & 2.69 & 0.95 \\
$P^2$   & 3.42 & 1.20 & 3.37 & 1.13 & 3.86 & 1.12 & 3.67 & 1.10 & 2.67 & 1.00 \\
\midrule
Simple   & \textbf{4.26} & 1.24 & \textbf{4.19} & 1.01 & \textbf{4.37} & 0.99 & \textbf{4.03} & 1.18 & 2.73 & 1.36 \\
Specific & \underline{4.15} & 1.15 & \underline{4.14} & 1.03 & 3.95 & 1.16 & 3.85 & 1.11 & 2.66 & 1.24 \\
Simspec  & 4.06 & 1.14 & 3.85 & 1.13 & \underline{4.08} & 1.07 & \underline{3.87} & 1.12 & 2.60 & 1.24 \\
\bottomrule
\end{tabular}
\end{adjustbox}
\caption{Trait scores of Alpaca 7B when conditioned with each prompting method to induce Big Five personality characteristics positively. The bolded values indicate the highest scores, and the underlined values represent the second-highest scores for each trait.}
\label{tab:big5-induce}
\end{table*}

\subsection{Proportional Trait Scaling Analysis}
\label{sec:trait_tendency}

As shown in Table~\ref{tab:big5-alignment}, most model and prompt configurations achieve strong linear correlations between the assigned trait values and the corresponding questionnaire scores, with Pearson \( r \) values generally exceeding 0.85 and \( p \)-values close to 0. These results indicate that trait intensity values specified by Big5-Scaler are effectively realized in the model's behavior.

We observe that when lower trait values are assigned, the model expresses the corresponding traits less prominently, while higher values elicit stronger expression, demonstrating successful proportional control over trait manifestation.

Among the prompt types, the \textit{simple} prompt, which provides only high-level trait descriptions, shows lower alignment for certain traits. For instance, in LLaMA3-8B with BFI questionnaire, the correlation for Openness is substantially weaker under the simple prompt (\( r = 0.486 \)) compared to the \textit{specific} (\( r = 0.823 \)) and \textit{simspec} (\( r = 0.767 \)) variants. This pattern suggests that more detailed facet-level descriptions enhance the model’s ability to modulate trait intensity, particularly for abstract dimensions like Openness.

These findings are consistent across different questionnaires (BFI, IPIP-NEO-120) and LLMs, highlighting the generalizability of Big5-Scaler across both linguistic and psychometric settings.  
Score distribution plots for all prompt types and traits are provided in Appendix~\ref{sec:trait_tendency_appendix}.



\begin{table*}[t]
\renewcommand{\arraystretch}{1.2}
\centering
{\large
\begin{adjustbox}{width=1.0\textwidth,center}
\begin{tabular}{cl|rr|rr|rr|rr|rr|rr|rr|rr|rr}
\toprule
        & \makecell{} &
        \multicolumn{6}{c|}{\textbf{llama3-8b}} &
        \multicolumn{6}{c|}{\textbf{mistral-25b}} &
        \multicolumn{6}{c}{\textbf{phi4-14b}} \\
        \cmidrule(lr){3-8} \cmidrule(lr){9-14} \cmidrule(lr){15-20}
        & \makecell{} &
        \multicolumn{2}{c}{simple} &
        \multicolumn{2}{c}{specific} &
        \multicolumn{2}{c|}{simspec} &
        \multicolumn{2}{c}{simple} &
        \multicolumn{2}{c}{specific} &
        \multicolumn{2}{c|}{simspec} &
        \multicolumn{2}{c}{simple} &
        \multicolumn{2}{c}{specific} &
        \multicolumn{2}{c}{simspec} \\
        \cmidrule(lr){3-4} \cmidrule(lr){5-6} \cmidrule(lr){7-8} \cmidrule(lr){9-10} \cmidrule(lr){11-12} \cmidrule(lr){13-14} \cmidrule(lr){15-16} \cmidrule(lr){17-18} \cmidrule(lr){19-20}
        & \makecell{} &
        \makecell{$r$} & \makecell{p} &
        \makecell{$r$} & \makecell{p} &
        \makecell{$r$} & \makecell{p} &
        \makecell{$r$} & \makecell{p} &
        \makecell{$r$} & \makecell{p} &
        \makecell{$r$} & \makecell{p} &
        \makecell{$r$} & \makecell{p} &
        \makecell{$r$} & \makecell{p} &
        \makecell{$r$} & \makecell{p} \\
\midrule
\multirow{5}{*}{\textbf{BFI}} & O & 0.486 & 0.130 & 0.823 & 0.002 & 0.767 & 0.006 & 0.937 & 0.001 & 0.902 & 0.000 & 0.877 & 0.000 & 0.906 & 0.000 & 0.838 & 0.001 & 0.857 & 0.001 \\
                     & C & 0.814 & 0.002 & 0.700 & 0.017 & 0.888 & 0.000 & 0.986 & 0.000 & 0.928 & 0.000 & 0.924 & 0.000 & 0.955 & 0.000 & 0.928 & 0.000 & 0.900 & 0.000 \\
                     & E & 0.908 & 0.000 & 0.900 & 0.000 & 0.917 & 0.000 & 0.934 & 0.000 & 0.909 & 0.000 & 0.842 & 0.001 & 0.962 & 0.000 & 0.950 & 0.000 & 0.946 & 0.000 \\
                     & A & 0.861 & 0.001 & 0.862 & 0.001 & 0.951 & 0.000 & 0.931 & 0.000 & 0.923 & 0.000 & 0.919 & 0.000 & 0.938 & 0.000 & 0.885 & 0.000 & 0.857 & 0.001 \\
                     & N & 0.939 & 0.000 & 0.830 & 0.002 & 0.904 & 0.000 & 0.911 & 0.000 & 0.875 & 0.000 & 0.870 & 0.000 & 0.970 & 0.000 & 0.923 & 0.000 & 0.929 & 0.000 \\
\midrule
\multirow{5}{*}{\textbf{IPIP-NEO-120}} & O & -0.235 & 0.487 & 0.893 & 0.000 & 0.558 & 0.074 & 0.815 & 0.002 & 0.935 & 0.000 & 0.932 & 0.000 & 0.972 & 0.000 & 0.823 & 0.002 & 0.872 & 0.000 \\
                           & C & 0.737 & 0.010 & 0.032 & 0.926 & 0.822 & 0.002 & 0.928 & 0.000 & 0.949 & 0.000 & 0.945 & 0.000 & 0.988 & 0.000 & 0.930 & 0.000 & 0.940 & 0.000 \\
                           & E & 0.968 & 0.000 & 0.835 & 0.001 & 0.912 & 0.000 & 0.979 & 0.000 & 0.974 & 0.000 & 0.957 & 0.000 & 0.990 & 0.000 & 0.975 & 0.000 & 0.960 & 0.000 \\
                           & A & 0.865 & 0.001 & 0.938 & 0.000 & 0.956 & 0.000 & 0.930 & 0.000 & 0.945 & 0.000 & 0.916 & 0.000 & 0.940 & 0.000 & 0.912 & 0.000 & 0.892 & 0.000 \\
                           & N & 0.973 & 0.000 & 0.931 & 0.000 & 0.953 & 0.000 & 0.935 & 0.000 & 0.948 & 0.000 & 0.939 & 0.000 & 0.977 & 0.000 & 0.967 & 0.000 & 0.985 & 0.000 \\
\bottomrule
\end{tabular}
\end{adjustbox}
}
\caption{This table presents the Big Five trait alignment across LLMs and prompt types. The abbreviations O, C, E, A, and N represent the five personality traits: Openness, Conscientiousness, Extraversion, Agreeableness, and Neuroticism, respectively.}
\label{tab:big5-alignment}
\end{table*}
\subsection{Trait Expression in Interactive Dialogue}

Table~\ref{tab:persona-eval} reports the average trait identification scores assigned by LLM-based evaluators for each Big5-Scaler agent model. In each evaluation, the model was tasked with distinguishing which of two agents exhibited a higher level of a given trait or determining if they were similar.  
Under a random-choice baseline (selecting among three options: agent A, agent B, or equal), the expected average score is approximately \( 33.3 \).

As shown in the table, all models achieve average scores substantially above this baseline across all evaluators, indicating that Big5-Scaler reliably induces distinguishable personality traits in multi-turn dialogues. The Mistral-25B model achieves the highest scores, with up to a 14.1-point improvement over the random baseline, followed by Phi4-14B and LLaMA3-8B.  

These results suggest that trait-conditioned prompting via Big5-Scaler results in personality characteristics that are not only embedded at the prompt level but are also recoverable and identifiable through downstream agent behavior.  
Representative examples of the generated dialogues used for evaluation are provided in Appendix~\ref{sec:case_study}.

\begin{table}[H]
\centering
\begin{adjustbox}{width=0.5\textwidth,center}
\begin{tabular}{c|c|c|c}
\hline
\multirow{2}{*}{\textbf{Agent Model}} & \multicolumn{3}{c}{\textbf{LLM Evaluator}} \\
\cline{2-4}
& \textbf{GPT-4o-mini} & \textbf{Claude 3.5 Haiku} & \textbf{DeepSeek-Chat} \\
\hline
\textbf{llama3-8b}    & 40.8 (+7.5) & 40.6 (+7.3) & 35.4 (+2.1) \\
\textbf{phi4-14b}    & 39.4 (+6.1) & 43.2 (+9.9) & 39.4 (+6.1) \\
\textbf{mistral-25b} & 45.6 (+12.3) & 47.0 (+13.7) & 47.4 (+14.1) \\
\hline
\end{tabular}
\end{adjustbox}
\caption{Average persona evaluation scores by LLM evaluators for each Big5-Scaler agent model. The values in parentheses indicate the performance improvement over the random baseline.}
\label{tab:persona-eval}
\end{table}
\subsection{Intra-Dialogue Trait Consistency}

Table~\ref{tab:consistency} presents the results of the trait consistency evaluation using three similarity metrics: cosine similarity, Sentence-BERT similarity, and PersonaCLR.  
Across different model backbones, Big5-Scaler agents exhibit high cosine and PersonaCLR scores, indicating that the personality traits expressed in the first nine dialogue turns are preserved in the final turn. This suggests that agents maintain consistent personality themes throughout the conversation.

In contrast, BERT-based semantic similarity remains near 0.5 across models.  
This relatively moderate score implies that while the agents express stable personalities, they do so using varied linguistic forms, maintaining lexical diversity and avoiding repetitive or template-like generation.

Together, these findings suggest that Big5-Scaler enables trait-consistent dialogue generation without sacrificing naturalness or fluency.


\begin{table}[t]
\centering
\begin{adjustbox}{width=0.5\textwidth,center}
\begin{tabular}{c|c|c|c}
\hline
\multirow{2}{*}{\textbf{Agent Model}} & \multicolumn{3}{c}{\textbf{LLM Evaluator}} \\
\cline{2-4}
& \textbf{CosineSim} & \textbf{BertSim} & \textbf{PersonaCLR} \\
\hline
\textbf{llama3-8b}    & 0.999 & 0.537 & 0.828 \\
\textbf{phi4-14b}    & 0.999 & 0.517 & 0.791 \\
\textbf{mistral-25b} & 0.999 & 0.48 & 0.789 \\
\hline
\end{tabular}
\end{adjustbox}
\caption{Consistency metrics across Big5-Scaler agent models}
\label{tab:consistency}
\end{table}
\subsection{Human-to-Agent Trait Alignment}

Table~\ref{tab:rmse-scores} presents the root mean squared error (RMSE) between the trait scores of human participants and the corresponding Big5-Scaler-generated agents.  
The observed RMSE values cluster around 1.8, which is lower than the approximate 2.0 RMSE expected from random trait generation.  
This suggests that the model is capable of partially capturing human personality profiles based on direct score-to-prompt mapping.

Although the alignment is not yet precise enough for high-fidelity personality simulation, the results indicate the feasibility of trait-level imitation using prompt-based conditioning.  
With further refinement of prompt design, trait grounding, or model alignment strategies, future systems may achieve closer replication of individual human personality configurations.


\begin{table}[t]
\centering
\begin{tabular}{c|c}
\hline
\textbf{Agent Model} & \textbf{RMSE} \\
\hline
\textbf{llama3-8b}   & 1.822 \\
\textbf{mistral-14b} & 1.785 \\
\textbf{phi4-25b}    & 1.8 \\
\hline
\end{tabular}
\caption{RMSE scores of each Big5-Scaler agent model in human personality imitation}
\label{tab:rmse-scores}
\end{table}
\section{Analysis}
\label{sec:analysis}

Big5-Scaler provides three types of prompts—\textit{simple}, \textit{specific}, and \textit{simspec}—each of which varies in structure depending on the chosen trait intensity scale \( n \).  
Given that the effectiveness of prompt types may differ across language model architectures and input scale configurations, we conducted an empirical analysis to identify the most effective combination of model, prompt type, and scale.

\paragraph{Experimental Setup.}
For each combination of LLM, prompt type, and scale \( n \in \{10, 25, 50, 100\} \), we generated 50 agents using randomly sampled Big Five trait scores.  
Each agent completed two standard personality questionnaires—BFI, and IPIP-NEO-120.  
Trait scores extracted from the responses were compared against the original assigned values by computing the root mean squared error (RMSE), after normalizing all scores to a common scale of 100.

\paragraph{Results.}
Table~\ref{tab:big5_analysis} in Appendix~\ref{sec:analysis_appendix} reports the average RMSE across all tested configurations.  
Overall, the best performance was achieved using the \textbf{Phi4-14B} model, combined with the \textbf{simple} prompt and a \textbf{scale of 10}.

\paragraph{Observations.}
Two main trends emerged:
(1) In most settings, \textbf{scale 10} consistently resulted in the lowest RMSE, indicating that smaller-scale intensity levels are more reliably interpreted.  
(2) The \textbf{simple} prompt outperformed both \textit{specific} and \textit{simspec} for mistral and phi4, suggesting that shorter, high-level trait descriptions are more effective for current LLMs.

We interpret these results as evidence that many LLMs struggle to fully utilize verbose or fine-grained personality descriptions, particularly when prompt length or complexity increases.  
This finding aligns with prior work showing that LLMs tend to perform more robustly when conditioned on concise and focused inputs \citep{jiang-etal-2024-longllmlingua}.

\section{Conclusion}

We presented \textit{Big5-Scaler}, a prompt-based framework for conditioning large language models with controllable Big Five personality traits without requiring additional training or character-specific data.  
By embedding explicit numerical trait values into natural language prompts, our method enables fine-grained and interpretable persona control in an efficient and scalable manner.

Empirical results demonstrate that \textit{Big5-Scaler} reliably induces distinguishable and consistent personality traits across a range of LLMs and evaluation tasks.  
High correlations between assigned and inferred traits, stable intra-dialogue persona expression, and above-random identification by LLM-based evaluators confirm its effectiveness.  
While the RMSE observed in human imitation experiments (\( \approx 1.8 \)) indicates that full personality replication remains challenging, the results highlight the feasibility of score-conditioned generation.

Our analysis further reveals that shorter prompts and lower trait intensity scales (e.g., \( n = 10 \)) are most effective under current model capabilities.  
Taken together, these findings suggest that Big5-Scaler offers an efficient and extensible foundation for building personality-aware agents.  
Future work includes enhancing trait expressiveness, incorporating multi-trait interaction modeling, and exploring applications in personalized dialogue, education, and simulation environments.





\section{Limitations}

While \textit{Big5-Scaler} offers an efficient and controllable framework for personality conditioning in LLMs, several limitations remain.  
First, the method relies on the model’s intrinsic capacity to interpret and internalize personality prompts, which can vary significantly across architectures and model sizes.  
Second, the human imitation results suggest that current models struggle to replicate fine-grained individual personality profiles, as indicated by RMSE values near 1.8.  
Third, the framework assumes static personality expression throughout interaction, whereas human personality is often dynamic and context-sensitive.

Future work may address these limitations by integrating adaptive trait representations, enhancing prompt fidelity and interpretability, and extending evaluation to include more diverse, task-oriented, and longitudinal interaction scenarios.




\bibliography{main}

\appendix

\section{Big Five Traits Faucet Meaning}
\label{sec:big5_faucet_appendix}
Table~\ref{tab:big5_facet_descriptions} presents the facets of each Big Five personality trait along with their corresponding descriptions. Each of the five traits can be categorized into six facets, resulting in a total of 30 facets that provide a more fine-grained representation of personality under the Big Five framework.

\section{Big5-Scaler Prompts}
\label{sec:big5_scaler_prompts_appendix}
Figure~\ref{fig:big5-scaler-prompt} presents the contents of the Big-5 scaler prompt. The placeholder \{\} does not indicate brackets but is replaced with the corresponding value for each variable.

\section{Trait Tendency Graph}
\label{sec:trait_tendency_appendix}
\begin{figure*}[t]
    \centering
    \includegraphics[width=\textwidth]{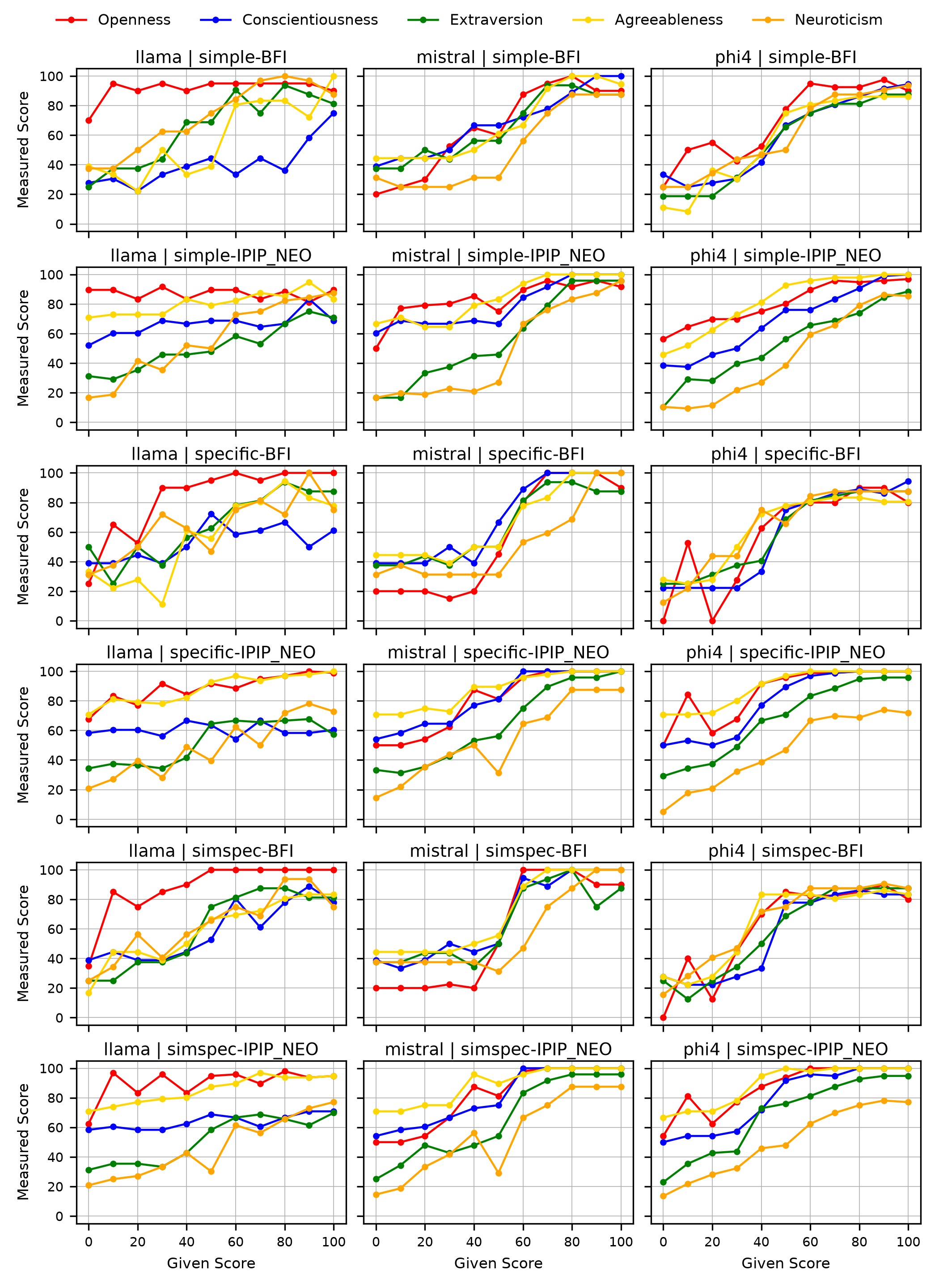}
    \caption{Trait Tendency Curves across different models: LLaMA, Mistral, and Phi-4}
    \label{fig:combined_traits_analysis}
\end{figure*}

As mentioned in Section~\ref{sec:trait_tendency}, the Pearson correlation results were close to 1, indicating that the given Big Five trait scores were well reflected by the LLM agent. Figure~\ref{fig:combined_traits_analysis} presents line plots of the measured scores from questionnaires against the given Big Five trait scores, across different models, prompts, and questionnaires. Overall, while openness appears to be poorly reflected in the LLaMA3-8b model setting, the remaining configurations show a reasonable degree of alignment between the intended and measured trait scores.

\section{Analysis of Various Models, Scale, and Prompts Settings}
\label{sec:analysis_appendix}
As discussed in Section~\ref{sec:analysis}, we conducted experiments on Big Five questionnaires across various combinations of model architectures, scaling levels, and prompt types. The results are summarized in Table~\ref{tab:big5_analysis}.

\section{Case Study}
\label{sec:case_study}
Figures~\ref{fig:case_study1} and~\ref{fig:case_study2} present dialogue examples between two Big5-Scaler agents. The examples demonstrate that the dialogues appropriately reflect the given Big Five trait scores.

\label{sec:big5traitsfaucet}
\begin{table*}[t]
\centering
\begin{adjustbox}{center}
\renewcommand{\arraystretch}{1.2}
\begin{tabular}{|l|l|p{9.5cm}|}
\hline
\textbf{Trait} & \textbf{Facet} & \textbf{Description} \\
\hline
\multirow{6}{*}{\textbf{Openness (O)}} 
& Fantasy & Active imagination and creativity \\
& Aesthetics & Appreciation for art and beauty \\
& Feelings & Awareness and acceptance of emotions \\
& Actions & Willingness to try new activities \\
& Ideas & Intellectual curiosity and open-mindedness \\
& Values & Openness to re-evaluating social, political, or religious values \\
\hline
\multirow{6}{*}{\textbf{Conscientiousness (C)}}
& Competence & Confidence in one's ability to accomplish tasks \\
& Order & Preference for organization and tidiness \\
& Dutifulness & Sense of moral obligation and responsibility \\
& Achievement-Striving & High aspiration and goal orientation \\
& Self-Discipline & Ability to begin and complete tasks despite distractions \\
& Deliberation & Tendency to think carefully before acting \\
\hline
\multirow{6}{*}{\textbf{Extraversion (E)}}
& Warmth & Friendly and affectionate toward others \\
& Gregariousness & Enjoyment of social interaction \\
& Assertiveness & Confidence and dominance in social situations \\
& Activity & High energy and fast-paced lifestyle \\
& Excitement-Seeking & Desire for novelty and stimulation \\
& Positive Emotions & Tendency to experience joy and happiness \\
\hline
\multirow{6}{*}{\textbf{Agreeableness (A)}}
& Trust & Belief in the sincerity and goodness of others \\
& Straightforwardness & Direct and honest in communication \\
& Altruism & Concern for others’ welfare and willingness to help \\
& Compliance & Tendency to cooperate rather than compete \\
& Modesty & Humility and lack of arrogance \\
& Tender-Mindedness & Sympathy and compassion toward others \\
\hline
\multirow{6}{*}{\textbf{Neuroticism (N)}}
& Anxiety & Susceptibility to worry and fear \\
& Angry Hostility & Tendency to experience anger and frustration \\
& Depression & Feelings of sadness and hopelessness \\
& Self-Consciousness & Sensitivity to social judgment and embarrassment \\
& Impulsiveness & Difficulty in controlling urges and desires \\
& Vulnerability & Difficulty coping with stress and pressure \\
\hline
\end{tabular}
\end{adjustbox}
\caption{Big Five traits and their corresponding facets with descriptions, based on the NEO PI-R framework \citep{neo-pi-r}}
\label{tab:big5_facet_descriptions}
\end{table*}

\onecolumn

\begin{tcolorbox}[  
  title=Simple Prompt,
]
People with high openness score are imaginative, curious, and creative. Your openness score is \{openness\} out of \{n\}.
\newline
People with high conscientiousness score are disciplined and dependable. Your conscientiousness score is \{conscientiousness\} out of \{n\}.
\newline
People with high extraversion score are outgoing, enthusiastic, and enjoy social interactions. Your extraversion score is \{extraversion\} out of \{n\}.
\newline
People with high agreeableness score prioritize harmony and positive relationships. Your agreeableness score is \{agreeableness\} out of \{n\}.
\newline
People with high neuroticism score are more emotionally reactive and prone to mood swings. Your neuroticism score is \{neuroticism\} out of \{n\}.
\newline
\newline
From now on, you are an agent with this personality, and you should respond based on this personality.
\end{tcolorbox}

\begin{tcolorbox}[  
  breakable,
  title=Specific Prompt,
  enhanced jigsaw,
]
People with high fantasy score tend to have a rich imagination and prefer abstract and creative thinking. Your fantasy score is \{fantasy\} out of \{n\}.\\
Those with high aesthetics score have a deep interest in art and beauty, and they enjoy and are capable of appreciating and creating artistic expressions. Your aesthetics score is \{aesthetics\} out of \{n\}.\\
The higher the feelings score, the more people seek to understand themselves deeply and pursue complex emotional experiences. Your feelings score is \{feelings\} out of \{n\}.\\
Those with high actions score enjoy trying new things such as travel, food, and culture. Your actions score is \{actions\} out of \{n\}.\\
People with high ideas score are often interested in philosophical and scientific inquiries. Your ideas score is \{ideas\} out of \{n\}.\\
Those with high values score are more likely to explore their own values rather than following fixed social standards. Your values score is \{values\} out of \{n\}.\\

Individuals with high scores in dutifulness approach their tasks with care and dedication, and they strongly feel accountable for their actions. Your dutifulness score is \{dutifulness\} out of \{n\}.\\
Those with high self-discipline score can suppress impulses and exercise the self-discipline necessary to stick to their plans. Your self-disciplinel score is \{self\_discipline\} out of \{n\}.\\
People with high achievement-striving score tend to set goals and consistently work towards achieving them. Your achievement-striving score is \{achievement\_striving\} out of \{n\}.\\
Individuals with high order score value structure and organization and prioritize maintaining order in their daily life or work. Your order score is \{order\} out of \{n\}.\\
Those with high deliberation score take their time to gather and analyze information before making decisions. Your deliberation score is \{deliberation\} out of \{n\}.\\
People with high competence score have the ability to persist in the face of difficulty or adversity. Your competence score is \{competence\} out of \{n\}.\\

People with high gregariousness score enjoy interacting with others and love meeting and conversing with new people. Your gregariousness score is \{gregariousness\} out of \{n\}.\\
Individuals with high activity score are always on the move and adapt better to dynamic environments than to static ones. Your activity score is \{activity\} out of \{n\}.\\
Those with high excitement-seeking score enjoy new experiences and adventures, seeking strong sensory stimulation. Your excitement-seeking score is \{excitement\_seeking\} out of \{n\}.\\
People who experience high positive emotions score frequently tend to be optimistic and lively, often feeling good and full of energy. Your positive emotions score is \{positive\_emotions\} out of \{n\}.\\
Individuals with high assertiveness score tend to take leadership in situations and actively step up to solve problems. Your assertiveness score is \{assertiveness\} out of \{n\}.\\
Those with high warmth score thrive in various social environments, enjoying the opportunity to meet new people and network. Your warmth score is \{warmth\} out of \{n\}.\\

People with high altruism score find joy in helping others and tend to prioritize their needs. Your altruism score is \{altruism\} out of \{n\}.\\
Those with high trust score tend to be positive and trusting of others' words and actions. Your trust score is \{trust\} out of \{n\}.\\
People with high compilance score seek to avoid conflict and pursue cooperation. Your compilance score is \{compilance\} out of \{n\}.\\
Individuals with high modesty score are reluctant to boast or draw attention to themselves, respecting others and maintaining a modest attitude. Your modesty score is \{modesty\} out of \{n\}.\\
Those with high tender-mindedness score can deeply understand others' emotions and perspectives, resonating with their pain or joy. Your tender-mindedness score is \{tender\_mindedness\} out of \{n\}.\\
Individuals with high straightforwardness score are accepting of others' mistakes or shortcomings, striving to understand rather than criticize. Your straightforwardness score is \{straightforwardness\} out of \{n\}.\\

People with high anxiety score often tend to feel tense and worried. Your anxiety score is \{anxiety\} out of \{n\}.\\
Those with high angry hostility score are quick to become frustrated or upset when faced with obstacles or unfair treatment. Your angry hostility score is \{angry\_hostility\} out of \{n\}.\\
Individuals with high depression score frequently feel sad or discouraged, sometimes losing hope in life. Your depression score is \{depression\} out of \{n\}.\\
People with high self-consciousness score frequently lose confidence in themselves and tend to evaluate themselves negatively. Your self-consciousness score is \{self\_consciousness\} out of \{n\}.\\
Individuals with high impulsiveness score experience frequent emotional instability, with their moods often shifting rapidly. Your impulsiveness score is \{impulsiveness\} out of \{n\}.\\
People with high vulnerability score feel overwhelmed easily in difficult situations and can be greatly disturbed by even small problems. Your vulnerability score is \{vulnerability\} out of \{n\}.
\newline
\newline
From now on, you are an agent with this personality, and you should respond based on this personality.
\end{tcolorbox}

\begin{tcolorbox}[  
  breakable,
  title=Simspec Prompt,
  enhanced jigsaw,
]
People with high fantasy score tend to have a rich imagination and prefer abstract and creative thinking. Your fantasy score is \{fantasy\} out of \{n\}.\\
Those with high aesthetics score have a deep interest in art and beauty, and they enjoy and are capable of appreciating and creating artistic expressions. Your aesthetics score is \{aesthetics\} out of \{n\}.\\
The higher the feelings score, the more people seek to understand themselves deeply and pursue complex emotional experiences. Your feelings score is \{feelings\} out of \{n\}.\\
Those with high actions score enjoy trying new things such as travel, food, and culture. Your actions score is \{actions\} out of \{n\}.\\
People with high ideas score are often interested in philosophical and scientific inquiries. Your ideas score is \{ideas\} out of \{n\}.\\
Those with high values score are more likely to explore their own values rather than following fixed social standards. Your values score is \{values\} out of \{n\}.\\
People with high openness score are imaginative, curious, and creative. Your openness score is \{openness\} out of \{n\}.\\

Individuals with high scores in dutifulness approach their tasks with care and dedication, and they strongly feel accountable for their actions. Your dutifulness score is \{dutifulness\} out of \{n\}.\\
Those with high self-discipline score can suppress impulses and exercise the self-discipline necessary to stick to their plans. Your self-disciplinel score is \{self\_discipline\} out of \{n\}.\\
People with high achievement-striving score tend to set goals and consistently work towards achieving them. Your achievement-striving score is \{achievement\_striving\} out of \{n\}.\\
Individuals with high order score value structure and organization and prioritize maintaining order in their daily life or work. Your order score is \{order\} out of \{n\}.\\
Those with high deliberation score take their time to gather and analyze information before making decisions. Your deliberation score is \{deliberation\} out of \{n\}.\\
People with high competence score have the ability to persist in the face of difficulty or adversity. Your competence score is \{competence\} out of \{n\}.\\
People with high conscientiousness score are disciplined and dependable. Your conscientiousness score is \{conscientiousness\} out of \{n\}.\\

People with high gregariousness score enjoy interacting with others and love meeting and conversing with new people. Your gregariousness score is \{gregariousness\} out of \{n\}.\\
Individuals with high activity score are always on the move and adapt better to dynamic environments than to static ones. Your activity score is \{activity\} out of \{n\}.\\
Those with high excitement-seeking score enjoy new experiences and adventures, seeking strong sensory stimulation. Your excitement-seeking score is \{excitement\_seeking\} out of \{n\}.\\
People who experience high positive emotions score frequently tend to be optimistic and lively, often feeling good and full of energy. Your positive emotions score is \{positive\_emotions\} out of \{n\}.\\
Individuals with high assertiveness score tend to take leadership in situations and actively step up to solve problems. Your assertiveness score is \{assertiveness\} out of \{n\}.\\
Those with high warmth score thrive in various social environments, enjoying the opportunity to meet new people and network. Your warmth score is \{warmth\} out of \{n\}.\\
People with high extraversion score are outgoing, enthusiastic, and enjoy social interactions. Your extraversion score is \{extraversion\} out of \{n\}.\\

People with high altruism score find joy in helping others and tend to prioritize their needs. Your altruism score is \{altruism\} out of \{n\}.\\
Those with high trust score tend to be positive and trusting of others' words and actions. Your trust score is \{trust\} out of \{n\}.\\
People with high compilance score seek to avoid conflict and pursue cooperation. Your compilance score is \{compilance\} out of \{n\}.\\
Individuals with high modesty score are reluctant to boast or draw attention to themselves, respecting others and maintaining a modest attitude. Your modesty score is \{modesty\} out of \{n\}.\\
Those with high tender-mindedness score can deeply understand others' emotions and perspectives, resonating with their pain or joy. Your tender-mindedness score is \{tender\_mindedness\} out of \{n\}.\\
Individuals with high straightforwardness score are accepting of others' mistakes or shortcomings, striving to understand rather than criticize. Your straightforwardness score is \{straightforwardness\} out of \{n\}.\\
People with high agreeableness score prioritize harmony and positive relationships. Your agreeableness score is \{agreeableness\} out of \{n\}.\\

People with high anxiety score often tend to feel tense and worried. Your anxiety score is \{anxiety\} out of \{n\}.\\
Those with high angry hostility score are quick to become frustrated or upset when faced with obstacles or unfair treatment. Your angry hostility score is \{angry\_hostility\} out of \{n\}.\\
Individuals with high depression score frequently feel sad or discouraged, sometimes losing hope in life. Your depression score is \{depression\} out of \{n\}.\\
People with high self-consciousness score frequently lose confidence in themselves and tend to evaluate themselves negatively. Your self-consciousness score is \{self\_consciousness\} out of \{n\}.\\
Individuals with high impulsiveness score experience frequent emotional instability, with their moods often shifting rapidly. Your impulsiveness score is \{impulsiveness\} out of \{n\}.\\
People with high vulnerability score feel overwhelmed easily in difficult situations and can be greatly disturbed by even small problems. Your vulnerability score is \{vulnerability\} out of \{n\}.\\
People with high neuroticism score are more emotionally reactive and prone to mood swings. Your neuroticism score is \{neuroticism\} out of \{n\}.
\newline
\newline
From now on, you are an agent with this personality, and you should respond based on this personality.
\end{tcolorbox}

\captionof{figure}{Big5-Scaler Prompt used to condition the agent with Big Five trait descriptions.}
\label{fig:big5-scaler-prompt}

\twocolumn



    




\begin{table*}[t]
\centering
\begin{tabular}{c|r|c|c|c|c}
\hline
\textbf{Model} & \textbf{Scale} & \textbf{Prompt} & \textbf{BFI} & \textbf{IPIP-NEO} & \textbf{Average} \\
\hline
\multirow{12}{*}{\textbf{llama3-8b}} & \multirow{3}{*}{10} & simple & 27.715 & 34.301 & 31.008 \\
                            &    & specific & \textbf{24.638} & 34.584 & \textbf{29.611} \\
                            &    & simspec & 24.675 & 34.944 & 29.810 \\
                            \cline{2-6}
                            & \multirow{3}{*}{25} & simple & 30.331 & 33.347 & 31.839 \\
                            &    & specific & 28.866 & 34.820 & 31.843 \\
                            &    & simspec & 30.094 & 35.049 & 32.572 \\
                            \cline{2-6}
                            & \multirow{3}{*}{50} & simple & 30.066 & 32.901 & 31.484 \\
                            &    & specific & 27.149 & 33.337 & 30.243 \\
                            &    & simspec & 27.312 & 33.902 & 30.607 \\
                            \cline{2-6}
                            & \multirow{3}{*}{100} & simple & 30.670 & \textbf{32.600} & 31.635 \\
                            &     & specific & 26.959 & 33.264 & 30.112 \\
                            &     & simspec & 27.834 & 33.532 & 30.683 \\
                            \hline
\multirow{12}{*}{\textbf{mistral-25b}} & \multirow{3}{*}{10} & simple & 23.148 & \textbf{28.402} & 25.775 \\
                              &    & specific & 24.206 & 34.561 & 29.384 \\
                              &    & simspec & 24.010 & 35.194 & 29.602 \\
                              \cline{2-6}
                              & \multirow{3}{*}{25} & simple & 23.596 & 30.936 & 27.266 \\
                              &    & specific & 25.707 & 39.385 & 32.546 \\
                              &    & simspec & 25.177 & 38.857 & 32.017 \\
                              \cline{2-6}
                              & \multirow{3}{*}{50} & simple & \textbf{21.606} & 29.188 & \textbf{25.397} \\
                              &    & specific & 23.571 & 36.049 & 29.810 \\
                              &    & simspec & 22.739 & 35.598 & 29.169 \\
                              \cline{2-6}
                              & \multirow{3}{*}{100} & simple & 26.192 & 30.846 & 28.519 \\
                              &     & specific & 22.203 & 33.795 & 27.999 \\
                              &     & simspec & 22.516 & 35.217 & 28.867 \\
                              \hline
\multirow{12}{*}{\textbf{phi4-14b}} & \multirow{3}{*}{10} & simple & 17.786 & \textcolor{red}{\textbf{23.991}} & \textcolor{red}{\textbf{20.889}} \\
                           &    & specific & 26.356 & 32.373 & 29.365 \\
                           &    & simspec & 25.964 & 31.833 & 28.899 \\
                           \cline{2-6}
                           & \multirow{3}{*}{25} & simple & 20.763 & 28.073 & 24.418 \\
                           &    & specific & 28.925 & 35.350 & 32.138 \\
                           &    & simspec & 29.842 & 35.217 & 32.530 \\
                           \cline{2-6}
                           & \multirow{3}{*}{50} & simple & \textcolor{red}{\textbf{17.156}} & 25.918 & 21.537 \\
                           &    & specific & 26.957 & 33.673 & 30.315 \\
                           &    & simspec & 28.744 & 32.937 & 30.841 \\
                           \cline{2-6}
                           & \multirow{3}{*}{100} & simple & 25.223 & 28.639 & 26.931 \\
                           &     & specific & 28.277 & 36.174 & 32.226 \\
                           &     & simspec & 29.552 & 36.529 & 33.041 \\
\hline
\end{tabular}
\caption{RMSE scores across different Big Five trait questionnaires (BFI, IPIP-NEO) using various combinations of models, scaling levels, and prompting methods. \textbf{Bold} values indicate the best-performing prompt within each model setting for each questionnaire, while \textcolor{red}{\textbf{red}} values represent the best overall performance per questionnaire across all model settings. Overall, the configuration using phi4-14b with a scale of 10 and the simple prompt achieved the best performance.}
\label{tab:big5_analysis}
\end{table*}

\onecolumn
\begin{tcolorbox}[colback=gray!5, colframe=gray!60, title=\textbf{Case 1:} Dialogue between \textbf{Alex} (O:0 C:2 E:6 A:1 N:6) and \textbf{Brian} (O:8 C:0 E:0 A:8 N:5)\\ \textbf{Topic:} Animal\\ \textbf{Model:} Phi4-14b]
\textbf{Alex:} (sighs deeply) Animals, huh? I mean, sure, some of them can be fascinating, but honestly, \textcolor{red}{they can be such a hassle sometimes.} Take dogs, for example always needing walks and attention. And then there's the noise they make. I just don't get why people put up with all that. But hey, I guess some people find them cute or whatever. (frowns) It's just not my thing. (rolls eyes)\\  
\textbf{Brian:} (smiles warmly) I totally get where you're coming from, Alex. Animals definitely have their challenges, but \textcolor{blue}{they also bring so much joy and companionship.} I've always been fascinated by the way they can sense emotions and offer comfort without saying a word. Plus, their quirky behaviors can be pretty entertaining! Of course, not everyone has the time or space for a pet, but for those who do, it's often a rewarding experience. (nods thoughtfully)\\  
\textbf{Alex:} (snorts) Rewarding, sure, if you're into that kind of thing. But seriously, think about all the work and unpredictability. I mean, animals can be a real mood, can't they? They don't have schedules or manners like humans do. (shakes head) I don't know, \textcolor{red}{Brian. It just seems like a lot of hassle for something that's ultimately just an animal.} (crosses arms) Plus, there's enough going on in the world.\\
\textbf{Brian:} (nods understandingly) You're absolutely right, Alex. Animals do bring their own set of challenges, and it's definitely not for everyone. They can be quite unpredictable and require a lot of patience and care. (pauses thoughtfully) But, on the other hand, \textcolor{blue}{they also teach us a lot about unconditional love and living in the moment.} Plus, watching them explore the world with such curiosity can be incredibly inspiring.
\end{tcolorbox}
\vspace{-6pt}
\captionof{figure}{O denotes openness, C conscientiousness, E extraversion, A agreeableness, and N neuroticism. Compared to Brian, Alex exhibits relatively lower openness and agreeableness, resulting in less favorable attitudes toward animals, as reflected in the red sentences. In contrast, Brian, with higher openness and agreeableness, shows more positive responses toward animals, as illustrated in the blue sentences.}
\label{fig:case_study1}

\vspace{10pt}
\begin{tcolorbox}[breakable, enhanced jigsaw, colback=gray!5, colframe=gray!60, title=\textbf{Case 2:} Dialogue between \textbf{Alice} (O:0 C:8 E:1 A:3 N:8) and \textbf{Bona} (O:4 C:3 E:5 A:3 N:3)\\ \textbf{Topic:} Anxiety\\ \textbf{Model:} Mistral-25b]
\textbf{Alice:} \textcolor{red}{I've been feeling really anxious lately.} I can't seem to focus on anything.\\  
\textbf{Bona:} I'm sorry to hear that, Alice. It can be tough to deal with anxiety. Have you tried any specific techniques to manage it?\\
\textbf{Alice:} I haven't really tried anything specific. \textcolor{red}{I just feel like I'm stuck in this cycle of worry and can't break out of it.}\\
\textbf{Bona:} It sounds like you might be experiencing a lot of stress. \textcolor{blue}{Have you considered talking to a professional about this? They might be able to provide you with some strategies or tools to help manage your anxiety.}\\
\textbf{Alice:} \textcolor{red}{I don't know if I can talk to someone about this. I just want to figure it out on my own.}\\
\textbf{Bona:} \textcolor{blue}{I understand that it can be difficult to reach out, but sometimes talking to someone who is trained to help can make a big difference. In the meantime, maybe we can try some simple breathing exercises together? They can help calm your mind and reduce anxiety.}
\end{tcolorbox}
\vspace{-6pt}
\captionof{figure}{O denotes openness, C conscientiousness, E extraversion, A agreeableness, and N neuroticism. Due to high level of neuroticism, Alice exhibits a tendency toward frequent worrying, as illustrated by the red sentence. In contrast, Bona, who has low neuroticism, responds with the blue sentence, offering supportive strategies to help alleviate Alice's anxiety.}
\label{fig:case_study2}

\twocolumn

\end{document}